# A novel Framework for Open-Vocabulary Multi-Object Recognition using CLIP

Wei Yu Chen
Faculty of Software and *Information Science*
*Iwate Prefectural University*
Takizawa, Japan
willychen850@gmail.com

Ying Dai
Faculty of Software and *Information Science*
*Iwate Prefectural University*
Takizawa, Japan
dai@iwate-pu.ac.jp

***Abstract*— To address the limitations of existing open-vocabulary object recognition methods, including high system complexity, substantial training costs, and limited generalization capability, this paper proposes a novel Open-Vocabulary Object Recognition (OVOR) framework based on a streamlined two-stage strategy: object segmentation followed by recognition. The proposed framework eliminates the need for complex retraining procedures and labor-intensive annotation. After extracting object regions, object-level image embeddings and category-level text embeddings are generated using CLIP, enabling recognition over arbitrary vocabularies. To reduce dependence on CLIP and enhance encoding flexibility, we further introduce a CNN/MLP-based approach that extracts convolutional neural network (CNN) feature maps and employs a multilayer perceptron (MLP) to align visual features with text embeddings. The resulting embeddings are then concatenated for cross-modal matching. Finally, object recognition is performed through similarity matching between image and text embeddings. Experiments on COCO, Pascal VOC, and ADE20K demonstrate that CLIP-based image encoding achieves the highest average AP, outperforming current state-of-the-art methods. Meanwhile, the results reveal the potential of CLIP-independent image encoding as a promising alternative for OVOR.**

## I. Introduction

Object recognition is fundamental to visual understanding and has advanced rapidly with deep learning and large-scale datasets, enabling applications such as autonomous driving, surveillance, industrial inspection, and HCI. However, most methods assume a closed set of training categories and often fail on unknown or novel classes. Because real-world categories constantly evolve, improving recognition beyond fixed labels is crucial. Open-vocabulary recognition addresses this by allowing arbitrary textual descriptions and enabling inference on unseen categories, making it practical when categories are various or exhaustive annotation is infeasible.

In recent years, with the rapid development of Vision–Language Models (VLMs), open-vocabulary object recognition has become an important research direction. Many studies attempt to exploit the cross-modal semantic space learned by models such as CLIP[7], introducing linguistic semantic knowledge into object detection and segmentation in order to overcome the limitations of traditional closed-set recognition. ViLD [1] distills CLIP/ALIGN but requires fine-tuning and complex distillation. MarvelOVD [2] uses CLIP pseudo labels and mining, yet is dataset-dependent and weak in transfer. Mask-adapted CLIP [3] improves ADE20K segmentation via mask prompts but may remove context and still needs retraining. HD-OVD [5] uses multi-level distillation with higher complexity. Distilled DETR [6] boosts COCO but adds branches/hyperparameters. In summary, existing open-vocabulary object recognition methods can improve performance on specific datasets, but they commonly suffer from complex training procedures, strong dataset dependency, and limited generalization. Therefore, how to effectively utilize the semantic capabilities of VLMs while reducing training cost and system complexity remains a promising and important research direction in open-vocabulary object recognition.

To address these issues, this study proposes a novel Open-Vocabulary Object Recognition (OVOR) framework based on a streamlined two-stage strategy: object segmentation followed by recognition. The proposed framework enables recognition without the constraints of a fixed category label set and can be implemented without complex retraining or labor-intensive annotation. The contributions of this paper are summarized as follows.

- Category-level text embeddings are generated via CLIP [7], which accommodates arbitrary vocabularies.
- After extracting object regions using an existing segmentation method [4], embeddings are generated using two distinct approaches: one utilizes the CLIP image encoder, while the other extracts CNN feature maps and employs a Multilayer Perceptron (MLP) to align visual features with text embeddings. The latter reduces reliance on CLIP and enhances encoding flexibility.
- Object and category embeddings are concatenated to construct a multimodal representation matrix. Object recognition is then performed through an embedding matching mechanism within this shared space.
- Experiments on three benchmark datasets, COCO, Pascal VOC, and ADE20K, demonstrate that CLIP-based

image encoding without Singular Value Decomposition (SVD) projection consistently achieves the highest average AP, outperforming representative state-of-the-art methods. In contrast, applying SVD projection yields higher Recall and Accuracy.

- Notably, CNN/MLP-based object-level image embeddings, generated independently of the CLIP image encoder, achieve viable recognition performance without the need for complex and expensive training.

## II. RELATED WORKS

### A. *Object segmentation*

Object segmentation is a key preprocessing step that separates candidate object regions for subsequent recognition. We adapt the unsupervised method in [4], which segments images using CNN-based EfficientNet features. It extracts pixel-level high-dimensional representations capturing local and global cues, applies PCA to emphasize dominant directions and suppress noise, and automatically determines the cluster number to adapt to image content. As it requires no human annotations and uses unsupervised clustering, it is well suited for open-vocabulary recognition.

### B. *Vision–language model*

Vision–language model (VLM) learns from images and text by mapping both modalities into a shared embedding space, enabling language-guided classification and zero-shot recognition. Representative VLMs include CLIP. CLIP uses a dual-encoder to embed images and text separately and match them by similarity. In this study we adopt CLIP as an encoder to get text embeddings of categories and image embeddings of object regions.

## III. METHODOLOGY

This study proposes a framework for open-vocabulary object recognition system, as shown in Fig. 1, which recognizes objects using arbitrary category names without relying on a fixed closed-set label space. Given an input image, an existing segmentation method is first applied to obtain candidate object regions [4]. The system then localizes each region to determine its position and spatial extent as the unit for feature extraction. Then, visual features are first extracted and aligned with text embeddings through an MLP-based image encoder. The image encoder generates embeddings for each object region, while the text encoder produces corresponding text embeddings for arbitrary category labels. The resulting image and text embeddings are concatenated and projected into a shared latent feature space, enabling unified similarity computation. Finally, each object region is assigned to the most relevant category based on embedding similarity matching. The system modules are described in detail as follows.

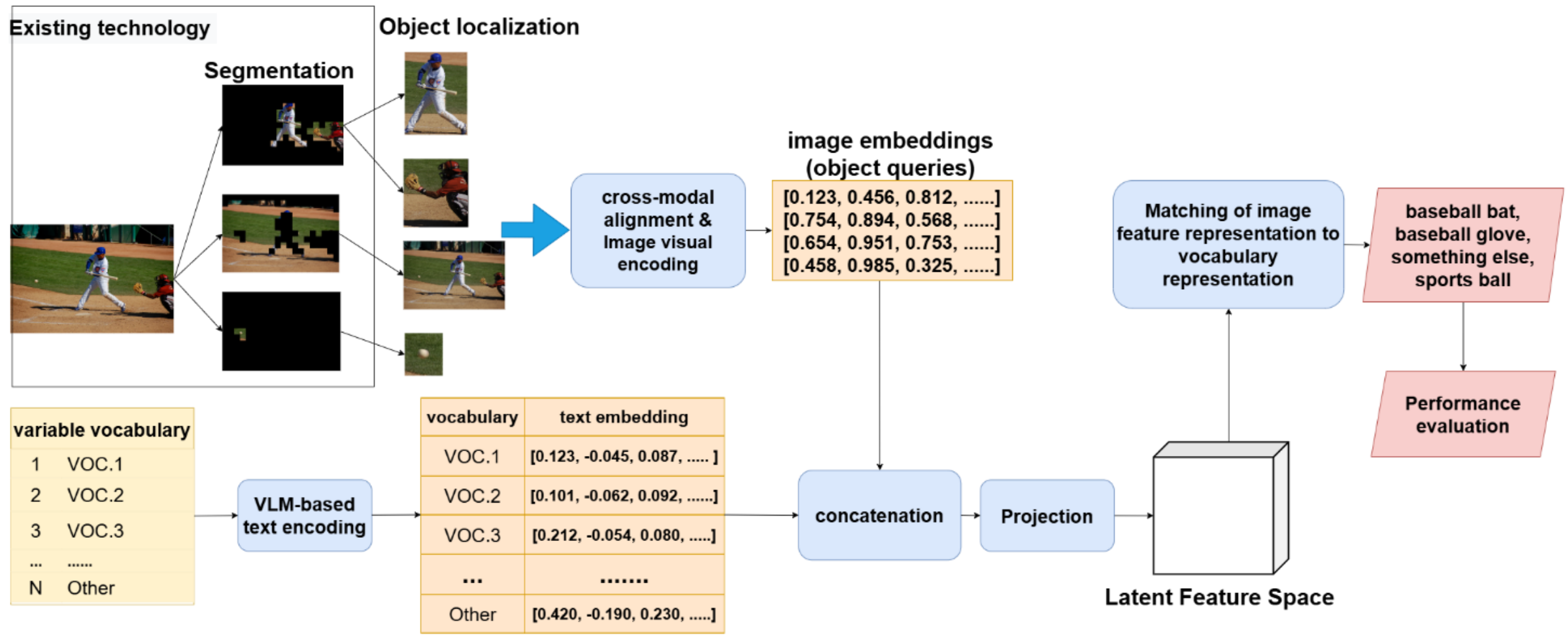

Fig. 1 Overview of open-vocabulary object recognition system

### A. *Object localization*

In our OVOR framework, cadidate regions of objects arefirstly locazized based on segmentatian results. The segment masks are resized to match the original image for pixel alignment. Connected-component analysis separates object regions, and small components are discarded as noise. A bounding box is computed for each valid region, and the corresponding patch is cropped from the original image for recognition.

### B. *Category-level text embedding*

For text embeddings, this study uses the CLIP text encoder (ViT-B/32) to convert category names into semantically meaningful vectors for matching. To support the open-vocabulary setting, we first define the target categories and their corresponding super-categories, and additionally include a "something else" category to handle instances that do not belong to the predefined set. We then design three prompt templates:

- Phrase1: a photo of a [super category] such as [category]
- Phrase2: this is a [category] of a [super category]
- Phrase3: a photo of [category]

All prompts are first tokenized into token sequences using the CLIP tokenizer and then fed into the CLIP text encoder to produce category-level text embeddings. Since the same CLIP model (ViT-B/32) is used on both the text and image sides, the resulting embeddings are ensured to lie in the same semantic space, facilitating subsequent similarity computation. To reduce the variance introduced by different prompt phrasings, we average the embeddings from Phrase1–Phrase3 to form an embedding called Avg Phrase.

Based on preliminary experiments, using Ave Phrase achieved the best object recognition performance. Therefore, in all subsequent experiments and quantitative evaluations, we use Avg Phrase as the text representation for candidate categories, and perform recognition during the matching stage by computing its similarity to the object-level image embeddings.

### *C. Object-level image embedding*

For image embeddings, we employ two methods to achieve visual feature representations, and analyze their effectiveness on open-vocabulary recognition performance.

The first approach directly uses the CLIP image encoder (ViT-B/32 backbone) to convert cropped object images into semantic embeddings. All images follow CLIP's standard preprocessing including resizing and normalization before being fed into the encoder. The resulting embeddings are L2-normalized onto the unit hypersphere. This approach corresponds to CLIP's native image–text semantic alignment and serves as the baseline of image representation.

To reduce reliance on CLIP and enhance encoding flexibility, the second approach uses a CNN-based feature extractor, EfficientNet-B0, to obtain feature maps from the input image. Specifically, we extract the head output feature map from EfficientNet-B0. Since those features are 2D spatial maps while CLIP text embeddings are 1D vectors with 512 dimensions, we further introduce an MLP to map the features into a fixed-dimensional image embedding space, enabling dimensional alignment between the image and the text modalities.

### *C. MLP-based image-text embedding alignment*

The MLP architecture is illustrated in Fig. 2. Its input is an array of visual feature maps produced by the EfficientNet-B0 head (7×7×1280). It consists of three fully connected layers with progressive dimensionality reduction, and the final output dimension is set to 512. This design matches the dimensionality of the text embeddings generated by the CLIP text encoder, enabling image and text embeddings to be directly matched within the same embedding space.

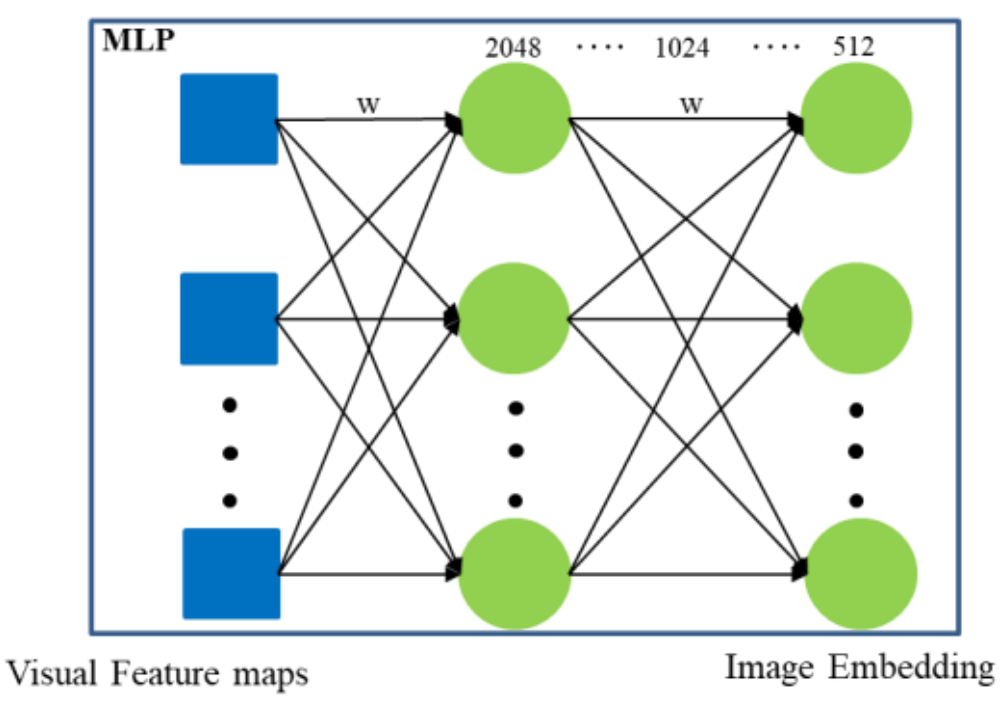

Fig. 2 Architecture of MLP

To train the MLP to generate image embeddings that can be aligned with the textual semantic space, this study adopts a distance-based loss function under a contrastive learning paradigm, as illustrated in Fig. 3. During training, for each sample we define an anchor (image embedding), a positive (the text embedding of the corresponding category), and a negative (the text embedding of a non-corresponding category). The loss function is defined as:

$$\text{Loss}=\max(0,d(a,p)-d(a,n)+\alpha) \quad (1)$$

where d( · ) denotes the distance between two embedding vectors, and α is a margin parameter that enforces a minimum separation between positive and negative pairs. With this loss, the MLP is optimized to reduce the distance between the image embedding and its matched text embedding while increasing the distance to text embeddings of mismatched categories, thereby progressively aligning the image embeddings to the text embedding space.

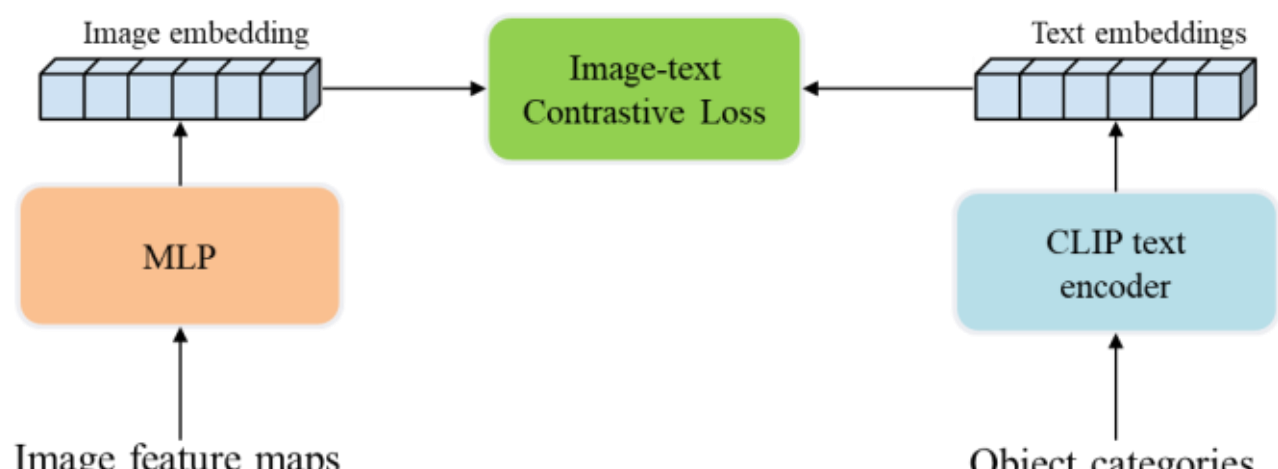

Fig. 3 Training pipeline of the MLP alignment model

### *D. Concatenation of embeddings*

To support similarity-based matching in a unified semantic space, we build an aligned image–text matrix via concatenation. For each input image, the system generates an embedding for every localized object region, then concatenates these object embeddings into an image-embedding matrix, where each row/column represents one object. In parallel, the corresponding text embeddings are concatenated in the same order to form a text-embedding matrix with identical structure. Then, these two matrixes are concatenated again to form a shared feature space. This one-to-one matrix correspondence enables later feature projection and cross-modal similarity computation.

### *E. Singular value decomposition-based projection*

To extract the principal characteristics and suppress noise, SVD is applied to the concatenated image and text embeddings to generate latent embeddings. First, Z-score standardization is performed so that each dimension has zero mean and unit variance; the resulting standardized matrix is denoted as Z. SVD is then applied to Z:

$$Z= U\Sigma V^T \quad (2)$$

where Z is the standardized embedding matrix; U is an orthogonal matrix whose columns are the left singular vectors; V is an orthogonal matrix whose columns are the right singular vectors; and Σ is a diagonal matrix containing the singular values.

By retaining the first k singular values and their corresponding singular vectors, the projection scores of Z in the latent space are computed as

$$Z_{SVD} = U_k\Sigma_k \quad (3)$$

where $Z_{SVD}$ denotes the matrix of k-dimensional latent embeddings.

Accordingly, SVD integrates image and text features into a shared latent space for subsequent matching and classification.

## F. *Matching and object recognition*

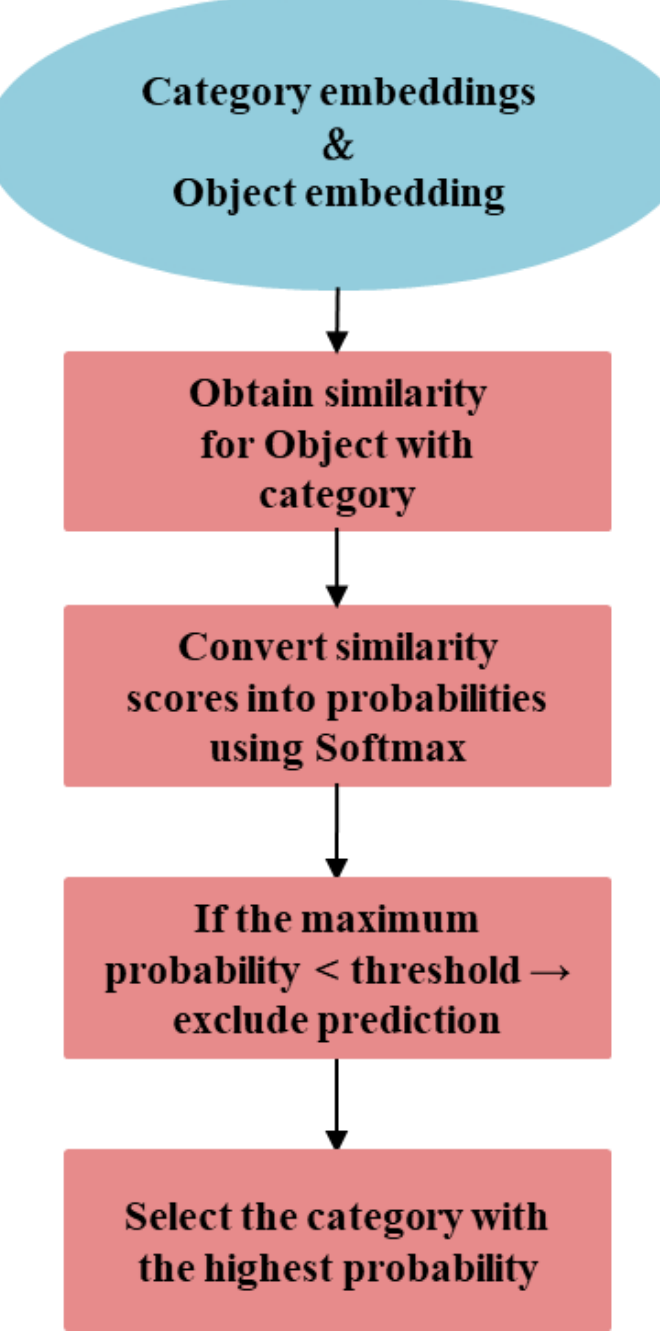

Fig. 4 Matching and object recognition

After alignment and concatenation, image and text embeddings are mapped into a shared feature space for matching and classification, as illustrated in Fig. 4. The system computes cosine similarity between each image embedding and all text embeddings, producing a similarity vector over candidate categories. These scores are then converted into a probability distribution using Softmax, which normalizes outputs to [0,1] and sums to 1. The category with the highest probability is selected as the recognition result. To reduce misclassification, a fixed threshold θ is applied after Softmax. If the maximum probability is below θ, the prediction is discarded. This filtering improves classification reliability and stability in object recognition.

# IV. EXPERIMENTAL RESULTS

## A. *Performance analysis of MLP*

We train the MLP with a contrastive learning strategy and analyze its behavior using training loss curves. Training data come from the ImageNet ILSVRC2012 validation set (50,000 images). Fig. 5 shows the loss trend over 0–200 epochs. The loss drops sharply in early epochs, indicating the MLP is learning to map visual features into the target embedding space. With extended training, the loss continues to decrease and converges to a lower range. Overall, the contrastively trained MLP reduces loss stably over time, and the gradual approach toward near-zero loss suggests increasing adaptation to the embedding-alignment objective.

For evaluation, we randomly sample 5,000 training images from ImageNet ILSVRC2012 for testing. We report class-wise and image-wise metrics (Precision, Recall, Accuracy, F1, and AP) to compare training stages. TABLE I, TABLE II, and TABLE III show results at Epoch 20, 200, and 2300. Metrics improve clearly from 20 to 200 epochs. With extended training to 2300 epochs, performance still increases but with smaller gains, indicating saturation. Thus, more iterations improve MLP recognition, but marginal benefits diminish after sufficient training.

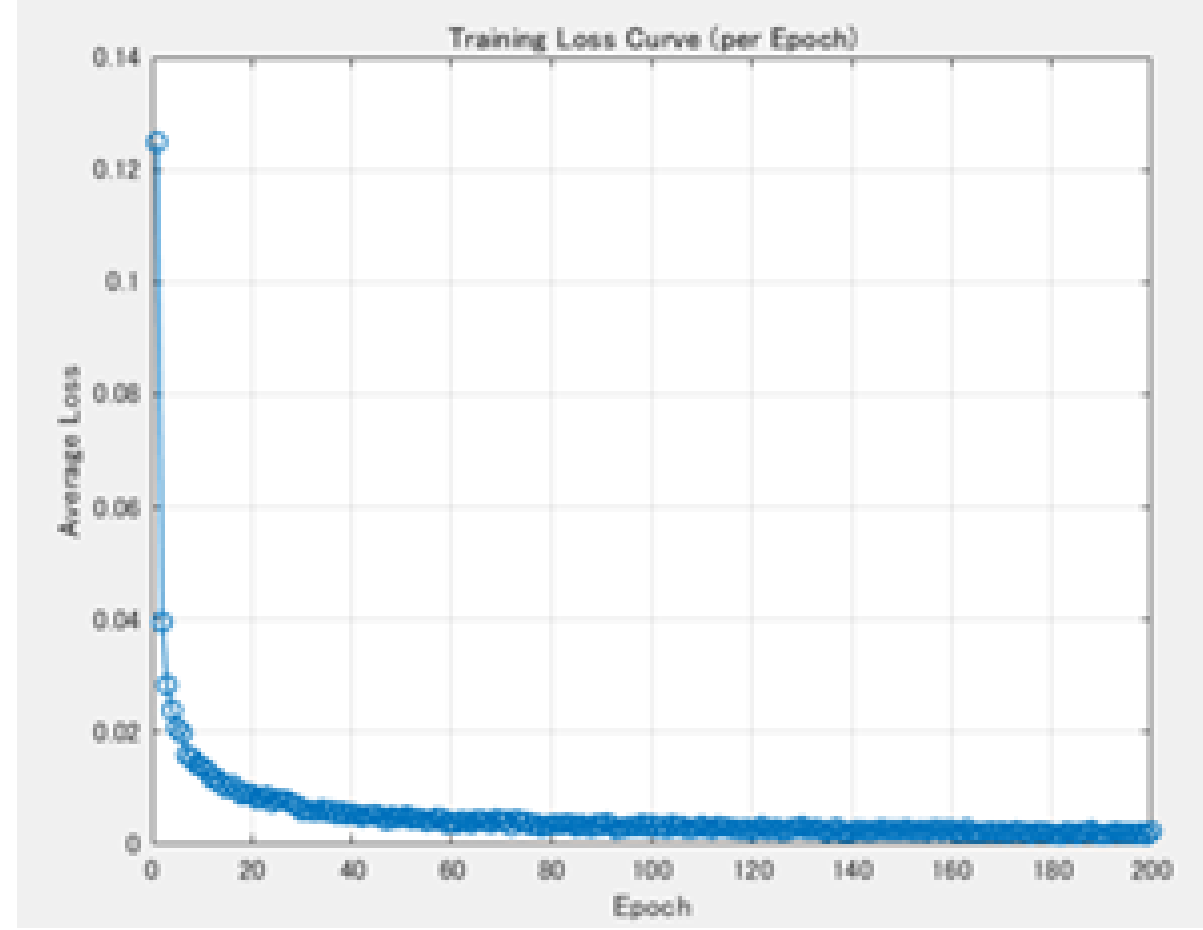

Fig. 5 Loss curves of MLP contrastive learning

TABLE I. PERFORMANCE OF MLP AT EPOCH 20

| Metric Type | Avg Precision | Avg Recall | Accuracy | F1 | AP |
|---|---|---|---|---|---|
| class-wise | 0.422 | 0.305 | 0.3 | 0.317 | 12.8 |
| image-wise | 0.303 | 0.303 | —— | 1 | 9.2 |

TABLE II. PERFORMANCE OF MLP AT EPOCH 200

| Metric Type | Avg Precision | Avg Recall | Accuracy | F1 | AP |
|---|---|---|---|---|---|
| class-wise | 0.606 | 0.493 | 0.495 | 0.468 | 29.8 |
| image-wise | 0.495 | 0.495 | —— | 1 | 24.5 |

TABLE III. PERFORMANCE OF MLP AT EPOCH 2300

| Metric Type | Avg Precision | Avg Recall | Accuracy | F1 | AP |
|---|---|---|---|---|---|
| class-wise | 0.646 | 0.581 | 0.583 | 0.577 | 37.5 |
| image-wise | 0.583 | 0.583 | —— | 1 | 34 |

## B. *Performance evaluation of object recognition on public dataset*

We evaluate object recognition performance on the public datasets COCO, Pascal VOC, and ADE20K. Category-level text embeddings are generated by the CLIP text encoder using the Avg Phrase strategy. For object-level image encoding, each image is segmented into candidate object regions using the method proposed in [4], and the resulting regions are encoded by both CLIP-based and MLP-based image encoders. The image and text embeddings are then concatenated for cross-modal matching. To investigate the effect of dimensionality reduction, SVD is applied to the concatenated embeddings. Since both the number of categories and the number of candidate object regions are smaller than the 512-dimensional embedding space, and the singular value spectrum of the shared region–category embedding space is relatively flat, the number of retained singular values, (k), is set equal to the number of categories in each dataset.

CLIP-based image encoding without SVD serves as the baseline for all comparisons.

Table IV presents the COCO results at a Softmax threshold of 0.0131, comparing CLIP-based and MLP-based image encoding methods with and without SVD. The value of $k$ is set to 80, corresponding to the number of categories in the COCO dataset, and 'Img Enc' denotes the image encoding method. The baseline achieves the highest AP at 41.9%. For the CLIP-based approach, applying SVD slightly improves Recall and Accuracy but significantly reduces Precision, F1-score, and AP. Regarding the CNN/MLP-based method (without SVD), the Accuracy reaches 37.8%, which is comparable to the 39.9% achieved by the CLIP-based method; however, other metrics show marked degradation, particularly Precision and AP.

TABLE IV. OBJECT RECOGNITION PERFORMANCE ON COCO

| Img Enc | SVD | Avg Precision | Avg Recall | Accuracy | F1 | AP |
|---|---|---|---|---|---|---|
| CLIP | X | 0.718 | 0.583 | 0.399 | 0.598 | 41.9 |
| CLIP | O | 0.449 | 0.617 | 0.481 | 0.456 | 27.7 |
| MLP | X | 0.357 | 0.467 | 0.378 | 0.344 | 16.7 |
| MLP | O | 0.193 | 0.133 | 0.084 | 0.095 | 2.6 |

TABLE V. OBJECT RECOGNITION PERFORMANCE ON VOC

| Img Enc | SVD | Avg Precision | Avg Recall | Accuracy | F1 | AP |
|---|---|---|---|---|---|---|
| CLIP | X | 0.959 | 0.757 | 0.689 | 0.794 | 72.6 |
| CLIP | O | 0.553 | 0.82 | 0.731 | 0.634 | 45.4 |
| MLP | X | 0.574 | 0.688 | 0.632 | 0.589 | 39.5 |
| MLP | O | 0.101 | 0.109 | 0.076 | 0.1 | 1.1 |

Table V presents the results for Pascal VOC at a Softmax threshold of 0.05, comparing CLIP- and MLP-based image encodings with and without SVD (*k=20)*. The baseline achieves the highest AP at 72.6%. When SVD is applied to CLIP, the AP decreases sharply, while Recall and Accuracy show slight improvements. For the CNN/MLP-based method (without SVD), Accuracy reaches 63.2%, which is 5.7% lower than the CLIP-based result; however, other metrics show significant degradation, particularly Precision and AP.

TABLE VI. OBJECT RECOGNITION PERFORMANCE ON ADE20K

| Img Enc | SVD | Avg Precision | Avg Recall | Accuracy | F1 | AP |
|---|---|---|---|---|---|---|
| CLIP | X | 0.347 | 0.367 | 0.246 | 0.271 | 12.7 |
| CLIP | O | 0.287 | 0.378 | 0.272 | 0.23 | 10.8 |
| MLP | X | 0.223 | 0.289 | 0.208 | 0.176 | 6.4 |
| MLP | O | 0.139 | 0.11 | 0.057 | 0.095 | 1.5 |

Table VI reports the ADE20K results at a Softmax threshold of 0.0085, comparing CLIP- and MLP-based image encodings with and without SVD (k = 120). The baseline achieves the highest AP of 12.7. When SVD is applied to CLIP, AP decreases slightly, while Recall and Accuracy increase marginally. For the CNN/MLP-based method (without SVD), Accuracy reaches 20.8%, which is 3.8% lower than the CLIP-based result; however, other metrics show marked degradation, particularly AP.

Results across the three datasets demonstrate that CLIP-based image encoding without SVD consistently achieves the highest AP. In contrast, SVD projection leads to higher Recall and Accuracy.

The CNN/MLP-based image embeddings consistently underperform CLIP-based embeddings, primarily due to insufficient cross-modal alignment. Mapping CNN features into the CLIP text space remains suboptimal, thereby reducing overall matching precision. Future work will focus on improving MLP alignment and image representations by: (1) retraining with the full ImageNet-1K dataset (~1.2M images); (2) defining a more effective loss function tailored for MLP training; and (3) incorporating objectives that preserve semantic structure and inter-class relationships.

Applying SVD to CLIP-based embeddings does not consistently improve performance. Because the singular-value spectrum of the image–text embedding matrix is relatively flat, the SVD-based projection may dilute category-discriminative information, thereby reducing the effectiveness of the latent representation for classification. Although Recall and Accuracy often increase after SVD, the accompanying decrease in Precision supports this observation, indicating a loss of class-discriminative capability in the projected embedding space. Moreover, the relatively flat singular-value spectrum of CLIP-generated image-text embeddings suggests that discriminative information is distributed across many dimensions, revealing a potential limitation of CLIP in producing highly category-discriminative embeddings.

For MLP-based encoding, the negative impact of SVD is more pronounced. The singular value spectrum of the image-text embeddings is heavily dominated by the first principal component, indicating that most information is concentrated in a single latent direction. Consequently, SVD-based projection leads to a substantial loss of discriminative information, resulting in degraded classification performance. These findings suggest that the current MLP-based alignment strategy remains insufficient for learning robust and discriminative image-text representations.

However, based on the above experimental results, we observe that when the MLP is trained and evaluated on the same dataset (ImageNet, 1,000 classes), the average recall and F1 score of object recognition using MLP-based image embeddings are comparable to those obtained with CLIP-based image embeddings on the COCO dataset (80 classes), while the classification accuracy is even higher. This finding suggests that, as a pretrained model, the MLP can be fine-tuned on a target dataset such as COCO to better adapt to its data distribution, thereby improving object recognition performance and potentially matching or even surpassing that achieved with CLIP-based image embeddings. In future work, we will implement and empirically validate this approach to examine its feasibility, aiming to develop an image encoding framework that does not rely on open-source pretrained models.

### *C. Comparison with SOTA methods*

TABLE VII. COMPARISON WITH SOTA METHODS

| Model | COCO-obj-80 | VOC | ADE20K |
|---|---|---|---|
| Extra training | | | |
| VilD[1] | 36.6 | 72.2 | —— |
| MarvelOVD[2] | 38.9 | —— | —— |
| HD-OVD[5] | 36.6 | —— | —— |
| DK-DETR[6] | 39.4 | 71.3 | —— |
| MaskCLIP[8] | —— | —— | 6.164 |
| OVOR(OURS,MLP) | 16.7 | 39.5 | 6.4 |
| training-free | | | |
| **OVOR(OURS,CLIP)** | **41.9** | **72.6** | **12.7** |

To validate effectiveness of the proposed framework, we compare our results with state-of-the-art (SOTA) methods The results are shown in TABLE VII. OVOR (OURS, CLIP) is training-free, using CLIP image encoding without SVD, while OVOR (OURS, MLP) uses MLP-based image encoding without SVD and requires extra training. OVOR (OURS, CLIP) achieves AP of 41.9 on COCO-obj-80, 72.6 on VOC, and 12.7 on ADE20K, outperforming the compared SOTA approaches. Unlike most prior methods that rely on extra-training or distillation, our proposed CLIP-based approach remains competitive without extra training.

An example of open-vocabulary object segmentation and recognition is presented in Fig. 6. The ground-truth (GT) categories in the image are {wine glass, dining table, fork, knife, person, pizza, TV}. Using CLIP-based image encoding without SVD, the recognized categories are {banana, person, pizza, wine glass, something else}. As illustrated in Fig. 6, salient objects, including the person and the pizza, are correctly recognized.

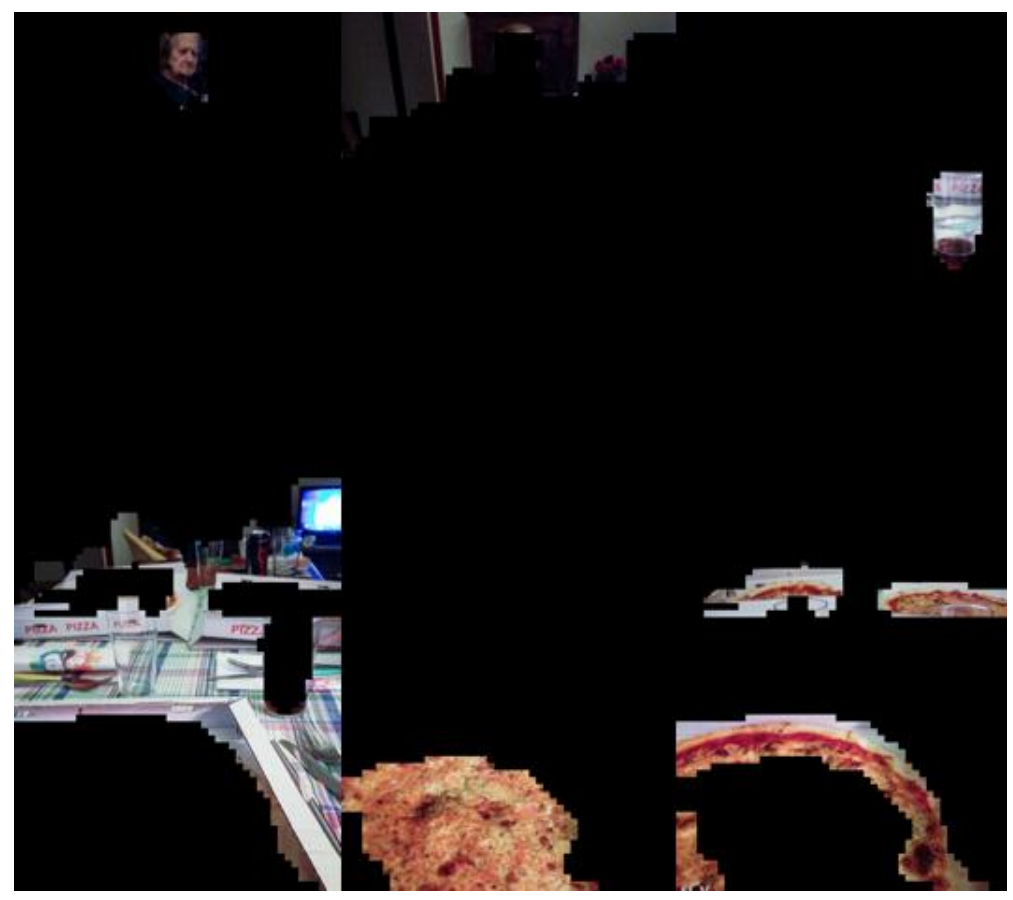

(a) Segmentation results

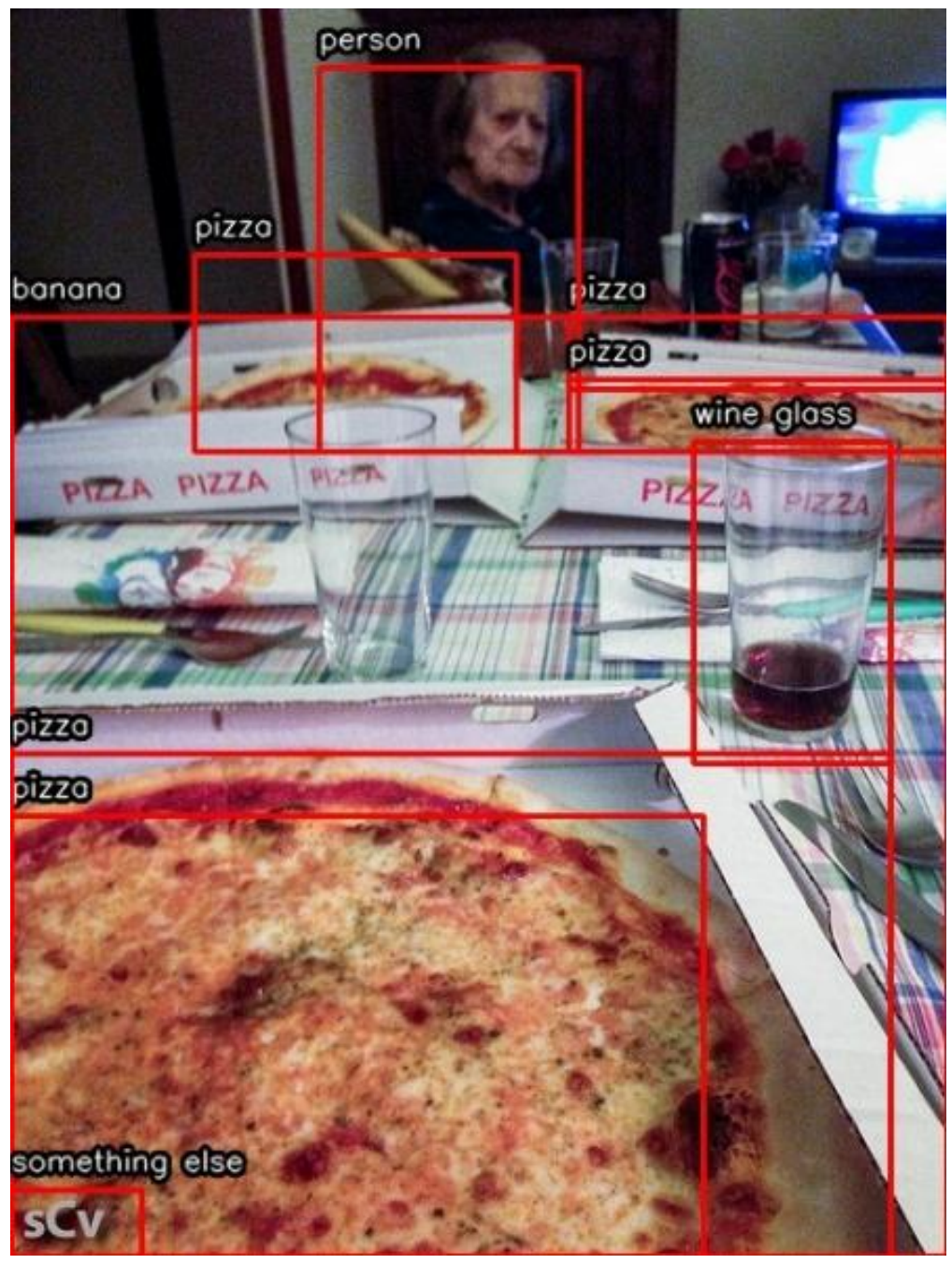

(b) Object recognition results

Fig. 6 An example of open-vocabulary object recognition

Overall, CLIP-based image encoding without SVD provides the most reliable performance, whereas SVD introduces a trade-off between Recall and Precision. While CNN/MLP-based image embedding alignment with CLIP-based text embeddings demonstrates the potential of CLIP-independent image encoding for OVOR, further improvements are needed to achieve competitive performance.

## V. CONCLUSION

We proposed a novel OVOR framework for recognizing diverse categories without relying on a fixed label set. The framework eliminates the need for complex retraining procedures and labor-intensive annotation.

Experiments on the COCO, Pascal VOC, and ADE20K datasets demonstrate that the training-free, CLIP-only configuration (without SVD) outperforms representative state-of-the-art methods. Furthermore, the results reveal the potential of CLIP-independent image encoding as a promising alternative for OVOR.

This work highlights both the strengths and limitations of CLIP representations. Our findings suggest that effective object-category alignment, rather than architectural complexity, is the key factor in developing low-cost open-world recognition systems.

Future work will focus on improving image-text alignment. We aim to develop more robust and flexible alignment strategies that enhance both visual and textual representations, thereby improving the generalization and practicality of open-vocabulary object recognition systems.

## ACKNOWLEAGEMENT

This work was supported in part by The Japan Society for the Promotion of Science (JSPS) KAKENHI under Grant JP22K12095, and in part by the Japan Keirin Autorace Foundation (JKA) Subsidy Program for Keirin and Auto Racing.